\documentclass[journal]{IEEEtran}
\usepackage{cite}

\ifCLASSINFOpdf
\else
\fi
\usepackage{amsmath}
\usepackage{graphicx}

\usepackage{array}
\usepackage{url}

\usepackage{amsmath, amssymb}

\begin{document}
%
\title{Understanding Character Recognition using Visual Explanations Derived from the Human Visual System and Deep Networks}
%
%
%

\author{Chetan Ralekar, Shubham Choudhary, 
        Tapan Kumar Gandhi*,
        Santanu Chaudhury 
\thanks{C. Ralekar, S. Choudhary, T.K.Gandhi, S. Chaudhury are with the Department
of Electrical Engineering, Indian Institute of Technology, New Delhi,
India, 110016 e-mail: chetan.ralekar@gmail.com}
\thanks{S. Chaudhury is currently associated with Department of Computer Science and Engineering, Indian Institute of Technology, Jodhpur, Rajastan}
\thanks{*corresponding author: tgandhi@iitd.ac.in}
\thanks{Manuscript received XXXXXXX}}

%
%

\markboth{The manuscript is submitted for review}%
{Shell \MakeLowercase{\textit{et al.}}: Bare Demo of IEEEtran.cls for IEEE Journals}
%



\maketitle

\begin{abstract}
Human observers engage in selective information uptake when classifying visual patterns. The same is true of deep neural networks, which currently constitute the best performing artificial vision systems. Our goal is to examine the congruence, or lack thereof, in the information-gathering strategies of the two systems. We have operationalized our investigation as a character recognition task. We have used eye-tracking to assay the spatial distribution of information hotspots for humans via fixation maps and an activation mapping technique for obtaining analogous distributions for deep networks through visualization maps. Qualitative comparison between visualization maps and fixation maps reveals an interesting correlate of congruence. The deep learning model considered similar regions in character, which humans have fixated in the case of correctly classified characters. On the other hand, when the focused regions are different for humans and deep nets, the characters are typically misclassified by the latter. Hence, we propose to use the visual fixation maps obtained from the eye-tracking experiment as a supervisory input to align the model’s focus on relevant character regions. We find that such supervision improves the model's performance significantly and does not require any additional parameters. This approach has the potential to find applications in diverse domains such as medical analysis and surveillance in which explainability helps to determine system fidelity.

\end{abstract}


%
\IEEEpeerreviewmaketitle

\section{Introduction}
%
%
%
%
\IEEEPARstart{W}{riting} symbols and characters are culture-specific, which involves arbitrary mapping between the visual shape and character identity \cite{NatureHuman}. Each writer has a distinct way of writing every character. Moreover, sophisticated design tools allow font designers to write a single character with a variety of artistic shapes and structures. As a result, we can find a single character with different visual representations having subtle variations in its structure or shape. For instance, a Latin character `A' can have many artistic variations in its shapes. 

Given such wide variability in a single character's visual appearance, correct identification depends on the information selection uptake from different character regions. In such scenarios, the visual system must be tuned to identify fine-grained details for visually similar characters\cite{NatureHuman} (such as F and E) and rely on coarser details in graphically dis-similar characters (like A and I ). Our sophisticated vision system uses such strategic information selection and processing, which enables us to discern discriminating and critical character regions. Using these key character regions, humans may easily recognize any deformed, artistic or handwritten characters. `Whether deep neural networks (DNNs) which constitute the leading artificial vision system, use similar information-gathering strategy?'  This is one of the interesting questions to be answered. 

The posed question can be addressed through deep investigations into the internal functioning of the DNNs. DNNs involve hierarchical non-linear mapping between the input and the output\cite{Yann}. Because of this mapping, the input-output relationship characterized by a DNN is highly complex, often including a large number of parameters, thereby making the exact functioning of the DNNs challenging to understand. To get some insights about DNNs, explainable techniques\cite{Simonyan,Deconvnet,shrikumar,Zintgra,bach2015pixel} attempt to provide human-understandable reasoning for the classifier's decision. Among these techniques, the visualization methods such as class activation mappings \cite{CAM}, and its variants \cite{GradCAM,GradCAM++} generate a heat map highlighting the key image regions that influence the model's decision. Such class activation mapping techniques are based on the intuition that the model observes some key pixels or regions of the image (those which provide high activations) while training. The model uses those regions to make a decision in favor of a particular class. Through visual inspection of these regions, humans can understand \textit{`where the model focuses during classification}'. These techniques can be useful while analyzing incorrectly classified samples, which can provide an opportunity to improve the model performance.

The above discussions emphasize the role key character regions play in the decisions made by humans and model in favor of a particular class. Such regions explaining the decision can be termed as `explanations'. After getting the explanations from both humans and models, it is worth exploring `whether the discriminating regions for humans and model are the same? If the key regions identified by the model are different than that of humans, is there a chance that the model misclassifies that particular sample? If so, can we improve the performance of the model using human explanations?' Let us try to shed some light on these exciting questions by utilizing the insights provided through explanations given by the deep network model and humans.

This paper is divided VI sections. Section I introduces the topic and section II highlights the related works. The details about the computational and human experimentation are given in section III. Section IV presents the experimental results which are followed by the discussions and future extensions in subsequent section. Concluding remarks are given in section VI.

\section{Related works}
In order to build trust in using deep learning architectures, researchers started working towards finding the proper reasoning or explanations for the network’s (model's) decision. The earlier efforts\cite{Simonyan,Deconvnet,shrikumar,Zintgra} attempt to understand network by doing sophisticated modifications such as occluding some regions or varying pixel intensities in the input image and then analyzing the response of the model. However, it becomes difficulty to do all possible modifications in the input data to understand the network completely. In order to find out which pixels contribute to the correct or wrong classification for each image separately, Sebastian Bach et al.\cite{bach2015pixel} have proposed a technique called Layer-Wise Relevance Propagation. In this method, the classification output is decomposed into sum of feature and pixel relevance score. For better visualization, a heat map is generated based on these relevance scores.

Other promising way of generating explanations is by highlighting the regions in the input image attended by the network while taking a decision, in favour of a particular class. Recently, the concept of class activation maps (CAM)\cite{CAM} have succeeded in localizing the discriminating images regions even when it has been trained for the task of classification. However, generation of CAM requires global average pooling layer to generate the maps which in turn demands architectural changes. To address this issue, gradient weighted class activation maps i.e. Grad-CAM \cite{GradCAM} is proposed. Grad-CAM\cite{GradCAM} and its variants\cite{GradCAM++} are applicable to wide variety of CNN family without any architectural modifications. All these techniques are trying to demonstrate why does the model's decision favour a particular class.

The state-of-the-art visualization techniques\cite{GradCAM,Deconvnet,CAM,GradCAM++,bach2015pixel} have made it possible to generate an explanation for classifier decision. Based on these maps, one can easily infer `why a particular sample (character) is misclassified?' However, these explanations are not compared with the explanations given by humans for the same task. Furthermore, these visualization techniques are only limited to generation of heat maps indicating image regions responsible for model's decision. Based on our knowledge, the use of these explanations for performance improvement of the model has been hardly attempted. Therefore, this paper not only proposes a modified visualization technique but also uses these explanations to design an attention mechanism for improving the model's performance. 

\section{Experimental Details}
We aim to get visual explanations from humans and DNNs. We will be discussing the experimental details regarding the same in this section. 
\subsection{Eye-tracking experiments}
Eyes are windows to perception and cognition which conveys wealth of information\cite{borji2014defending}. The eye movements can be considered to provide a valid measure of the spatial distribution of attention on account of the tight coupling between attention and eye movements in natural viewing tasks\cite{eyemovement_decision,borji2015reconciling}. Hence, eye-tracking has been used in various applications\cite{mao2021survey, takemura2014estimating, ralekar2017unlocking, ralekar2018effect, khanna2019memorability}. We are using eye-tracking methodology to unlock the mechanism of character recognition.    
\subsubsection{Stimulus Preparation}
\begin{figure*}[t]
    \centering
    \includegraphics[scale=0.4, trim={0 0cm 0 0}, clip]{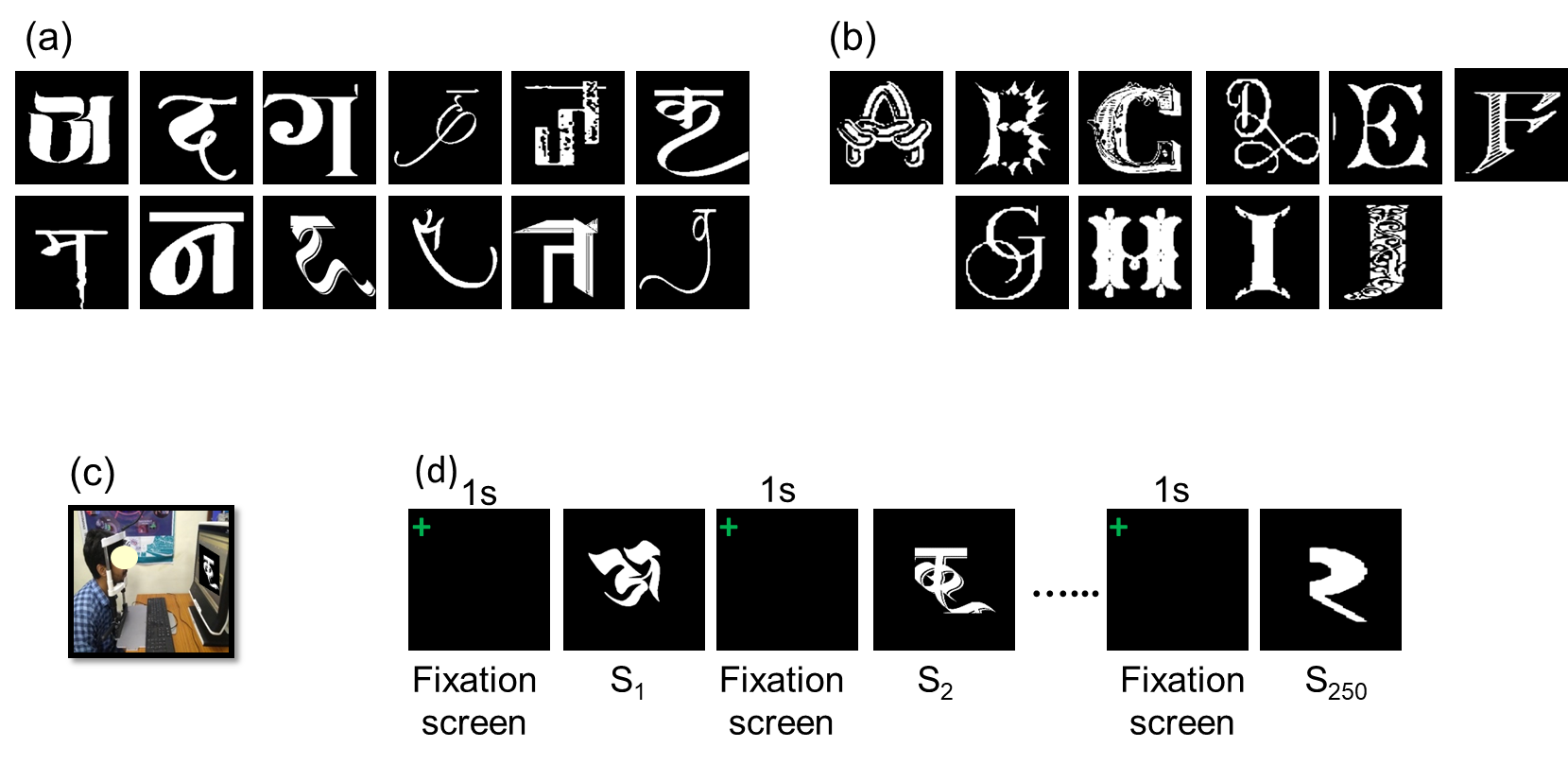}
    \caption{Randomly selected sample as a representative for each class of (a) Devanagari character and (b) Latin Character, (c) Participant sitting in front of the eye-tracker, (d) Stimulus presentation sequence.}
    \label{Fig:Ch3_Expt_protocol}
\end{figure*}

Devanagari script is used to write many official languages in India, and this script has a complex composition of its constituents symbols (i.e. curves, oriented edges)\cite{jayadevan2011offline}. We have chosen 12 characters from this script, formulating a twelve-class classification problem. Graphical structure\cite{bhagwat1961phonemic} , frequency of occurrence\cite{jayadevan2011offline}, and structural compositions\cite{puri2019efficient} are used as the selection criteria for these 12 classes of characters. We have used 1080 characters spread over 12 Devanagari character classes for the eye-tracking experiments. Besides, we have also created a dataset of 850 Latin (English) characters equally distributed in 10 classes. These ten classes are comprised of the first 10 characters from English (i.e. A, B, C, D, E, F, G, H, I, J). Both datasets are created by manually segmenting the characters from each word image downloaded from the internet and simple characters written with commonly used font styles. As a result, these datasets contain characters with various artistic variations in their structure and shapes and some characters in standard fonts such as Nirmala, Kokila for Devanagari, and Times New Roman, Aerial, etc., for Latin characters. We are having ground truth annotation associated with every character. All the characters are centered, resized to 400x400 pixels, and pre-processed to have the same background. The resultant samples can be seen in Fig. \ref{Fig:Ch3_Expt_protocol}(a) and Fig. \ref{Fig:Ch3_Expt_protocol}(b).

\subsubsection{Experimental protocol}
An eye-tracking experiment was conducted separately for Devanagari and Latin (English) characters. The characters from the respective datasets were randomly divided into different sets and were shown to 32 (17 males, 15 females) participants having normal or corrected to normal vision. The participants were graduate students of the Indian Institute of Technology, Delhi, with a mean age of $25 \pm2.5$ years, who could read Latin and Devanagari characters. All the participants were healthy and not suffering from any neurological disorders. Each character from the randomly chosen character set was presented to randomly chosen participants. Each character was presented on the screen of the Tobii T120 eye-tracker Fig. \ref{Fig:Ch3_Expt_protocol}(c) having a sampling frequency of 120 Hz. The distance between participant and eye-tracker was maintained around 60-70 cm. The participant was then asked to identify the character. It was a free-viewing task wherein participants could observe the stimulus image without any time restrictions. The stimulus remained on the screen till we got the response. Each trial consisted of a fixation screen followed by a stimulus (i.e. character) image. Each character image presentation was interleaved by the fixation cross at the extreme left of the screen presented for 1 second. The stimulus presentation sequence can be seen in Fig. \ref{Fig:Ch3_Expt_protocol}(d). The participant was expected to focus on the fixation-cross whenever it appeared on the screen. The location of fixation cross on extreme left avoids the issue of center bias \cite{Hartendorp}. The participants were supposed to respond by pressing the key and also read aloud the character identified so that experimenter could verify whether the response was correct or incorrect. To maintain a good quality of the eye-tracking data, the whole experiment was divided into 3 phases wherein each phase was preceded by a 9-point calibration process. Each phase comprised of around 250 trials. The participants took around 30 min to finish the experiment. 

\subsection{Eye-tracking data processing}
Eye-fixations are extracted from raw eye-gazes using Velocity-threshold (I-VT) algorithm\cite{salvucci2000identifying}. There was variability in recognition time among the participants. Some participants identified the character in less time, and some took more time to recognize it. This had resulted in variation in fixation duration. To counterbalance this effect, we had normalized the fixation duration using total time taken to recognize the particular character by each participant. These fixations corresponding to correct identification by several participants along with their normalized duration were combined and used to create the corresponding fixation map. Although the calibration step ensures minimized offset between estimated and actual gaze, there is an inherent error of around $1^{\circ}$ in the eye-tracker. This error is counter balanced by convolving a Gaussian mask over the collected fixations. We have used the method by O'Connel et. al. \cite{Oconnel} to generate the fixation maps using equation \ref{eq:Fixation_map}. The map is visualized as a heat map, and we name it as a fixation map.
\begin{equation}
\label{eq:Fixation_map}
F'(x,y) = \displaystyle\frac{1}{\sum\limits_{f=1}^{N_f} t_f} \sum\limits_{f=1}^{N_f} t_f \exp\left(\frac{-(x_f-x)^2 -(y_f-y)^2}{\sigma^2}\right)
\end{equation} 
where, $(x_f,y_f)$ are the fixation coordinates, $N_f$ is the number of fixations and  $t_f$ is corresponding normalized fixation duration, $\sigma=1^{\circ}$.

\subsection{Experiments using computational model}
\subsubsection{Dataset}
The character set used for the eye-tracking experiment is augmented by using shear (0.2), rotation ($\pm$40$^{\circ}$), vertical and horizontal flip to create a dataset consisting of 10800 and 8500 samples for Devanagari and Latin characters equally divided into 12 and 10 classes respectively. Out of those samples, 80\% samples are used for training, and 20\% are used for testing the model.

\subsubsection{Model Selection}
\begin{figure*}[t]
    \centering
    \includegraphics[scale=0.35, trim={0 0cm 0 0}, clip]{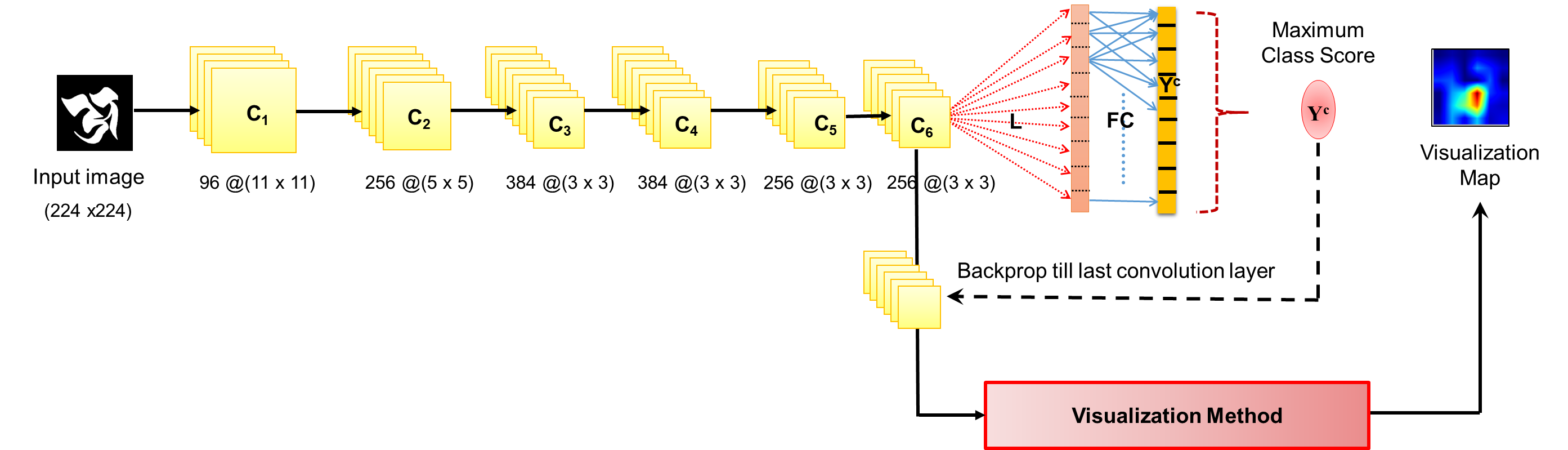}
    \caption{Baseline model with $C_1,C_2...C_6 :$ the convolutional layers, $L:$ the linearizeation (i.e. vectorization) of the map to 1D; $FC:$ the fully-connected layer.}
    \label{Fig:Ch3_Baseline_model}
\end{figure*}

The CNNs are the best feature extractor, and we are aiming to extract the best possible features. Therefore, we designed the model using only convolutional layers. The model consists of 6 convolutional layers as shown in Fig.\ref{Fig:Ch3_Baseline_model} followed by a single dense i.e., fully-connected layer acting as a linear classifier, and this forms our baseline model. The baseline model is inspired by AlexNet\cite{AlexNet} because it is suggested to provide the most brain-plausible object representations to humans and monkeys\cite{AlexNet,khaligh2014deep,fiset2009spatio}. Moreover, AlexNet has been used as a model of human vision in several studies\cite{pnas_pawan}. We have used a similar architecture as that of AlexNet\cite{AlexNet} with slight modification. We have added extra convolutional layers in our model and replaced all dense layers in AlexNet with a single dense layer at the output. We have used batch-normalization\cite{batchnorm} and dropout\cite{Dropout} with probability ($p = 0.2$) for all the convolutional layers. All the character images are resized to 224 x 224 before feeding it to the model.

\section{Results}
\subsection{Eye-fixations explain character recognition in humans}
It is worth noticing that humans do not process the visual scene at once. When humans observe an image or a scene, they tend to scan the image in several glimpses. With each spatial glance, new information is collected and iteratively processed\cite{Larochelle}, and the image is subsequently recognized. Such selective glimpses can be utilized to comment on the strategy used by humans for recognition. In the case of character recognition, the eye-gazes can help us get the key character regions responsible for character recognition. Moreover, the eye-glimpses i.e. eye-gazes, give an idea about moment-to-moment information processing\cite{rayner1998eye} that governs the observer's decision. Therefore, we have decided to collect eye-movement data for a recognition task hoping that capturing the eye-movement may enable us to understand the reasoning behind character identification. In other words, we can get some idea about `how a character is classified? which are the character regions that govern human's decision in favor of a character class?' We have designed and performed a simple recognition task wherein a participant is asked to identify a character displayed on the screen of an eye-tracker. In the course of recognition, the participants will direct their eye-gaze towards the information, which is of interest to them\cite{Salient}. These eye-gazes consist of quick, rapid eye-movements categorized as saccades and typically focused or sustained pauses of eyes called as fixations\cite{rayner1998eye}. During fixation, the eyes are relatively stable at a particular location over a certain period\cite{rayner1998eye}, and the brain processes visual information present at that location. On the other hand, the information processing is suppressed during saccade \cite{rayner1998eye}. 

\begin{figure}[t]
    \centering
    \includegraphics[scale=0.6, trim={0 0cm 0 0}, clip]{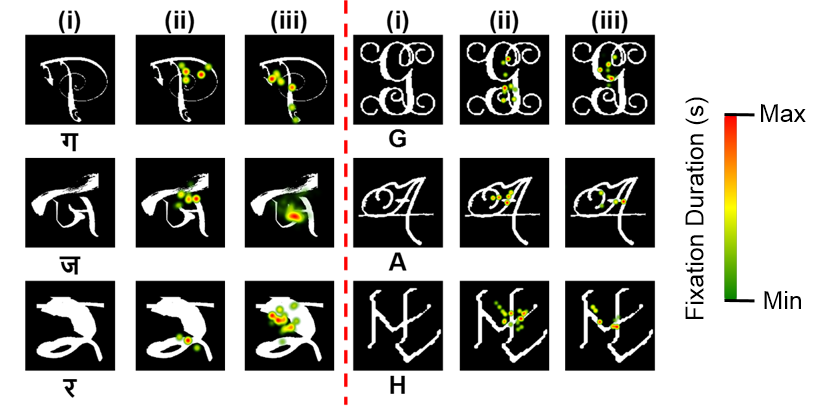}
    \caption{Fixations on different character regions reveal character identity. (i) Input character image, (ii) Fixation map for unrecognized character, (iii) Fixation map for correctly classified characters.}
    \label{Fig:Ch3_human_explanations}
\end{figure}

In an eye-tracking methodology, it is assumed that the attention of the decision maker is focused at the point of fixation\cite{eyemovement_decision}. The spatial distribution of such eye-fixations is an excellent indirect measure of the distribution of visual attention \cite{hoffman1998visual}. Several studies\cite{jiang2016learning,liu2016learning,mahdi2017comparison,mao2021survey} has established the importance of fixations in information processing. Hence, we have created the heat maps using eye-fixation information and named these maps as `fixation maps'. These maps allow us to know which part in the character image has influenced the participant's decision. It can be observed from Fig.\ref{Fig:Ch3_human_explanations}(iii), when a participant gives maximum attention to some specific regions of the character, he/she recognizes the character correctly. These particular regions might be contributing to the character identity and can be considered as key-character regions. On the contrary, when the participant cannot find the key regions and focuses on the artistic elements attached to the character or other character regions, the character is misclassified. For instance, participant's focus on extra-element, elongated part in character G as shown in Fig.\ref{Fig:Ch3_human_explanations}(ii) results in it's mis-classification. The selective attention on different character regions decides the participant's response in favor of the correct or incorrect class. Therefore, we may say that the spatial distribution of information hotspots inferred from fixation maps can explain the character recognition by humans to some extent.

\subsection{Generating explanations for model's decision}
To compare the information used by humans and model for recognizing a character, we need a map that is analogous to the fixation map. Generating the visualization maps for the classifier might help to comment on information selection uptake by the model. Among the various techniques of visualizing the model's decision \cite{Simonyan,Deconvnet,shrikumar,Zintgra,bach2015pixel,CAM}, the gradient weighted class activation mapping (Grad-CAM)\cite{GradCAM} is one of the most widely used methods for understanding the model's decision.  Grad-CAM provides a way to see which particular parts of the input image influenced the model's decision for a specifically assigned class label. This process gives some insight into the decision-making process of the model. Grad-CAM\cite{GradCAM} utilizes convolutional layers to generate maps because convolutional layers act as feature detectors. For instance, in an image classification task, the various layers of convolutional neural networks (CNNs) behave like the object detectors, even when there is no supervision on the object's location provided\cite{zhou2014object}. However, such ability to locate features is lost when we use fully-connected layers for classification\cite{CAM}. It implies that convolutional layers play a crucial role in generating the explanations of any CNNs. Because we are interested in getting discriminating character regions through model's explanations, we have selected a model that uses convolutional layers followed by a single dense layer as our baseline model.

\subsection{Visualization maps using Grad-CAM}
\begin{figure*}[t]
    \centering
    \includegraphics[scale=0.35, trim={0 0cm 0 0}, clip]{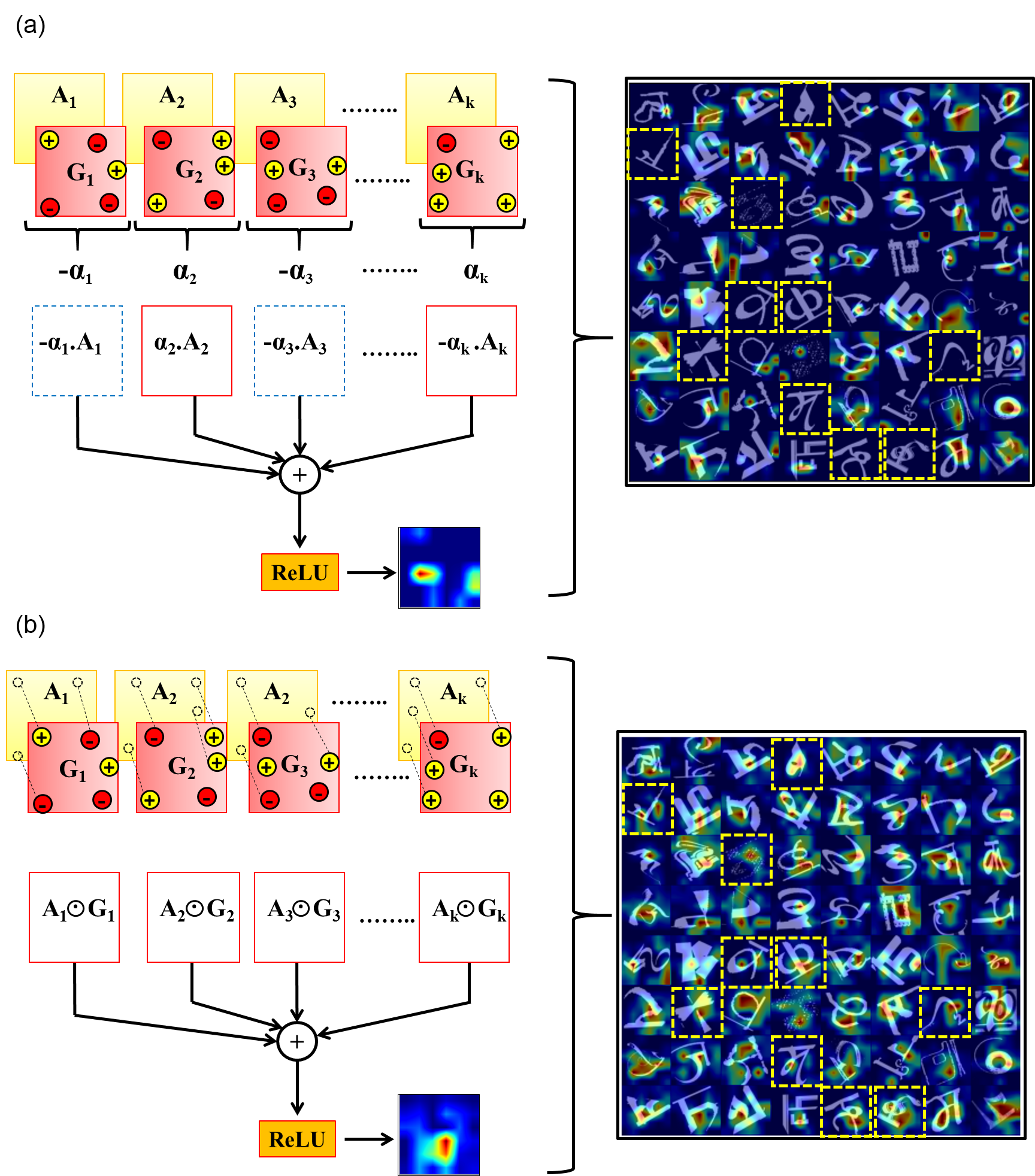}
    \caption{Generation of class activation maps (visualization maps) by (a) Conventional Grad-CAM\cite{GradCAM} and (b) Our method. The samples where Grad-CAM fails and our method generate the maps are highlighted by dotted yellow squares.}
    \label{Fig:Ch3_visualizations}
\end{figure*}

Given a CNN with a single dense layer as shown in Fig. \ref{Fig:Ch3_Baseline_model}, the algorithm of Grad-CAM\cite{GradCAM} starts with finding the gradient of the most dominant unit with respect to (w.r.t.) output activation map of the preceding convolution layer. Hence, we first compute the gradients of the score for the class $c$, $y^c$  w.r.t. output activation maps $\mathbf{A}^k$ of the penultimate activation layer (layer before the softmax) i.e.  $\frac{\partial y^c}{\partial A^k_{ij}}$. These gradients that are flowing backward are globally averaged pooled to obtain the neuron importance weight $\alpha^c_k$. This factor $\alpha^c_k$ captures the importance of feature map $k$ for a target (predicted) class $c$. The visualization maps are obtained as a weighted linear combination of forward activation maps, with $\alpha^c_k$ acting as the weight corresponding to the forward activation map $A^k$, followed by a ReLU. The process can be visualized from Fig. \ref{Fig:Ch3_visualizations}(a)

If $\mathbf{Y_{ij}^c}$ denote the ${ij}^{th}$ pixel for the activation map generated using the conventional Grad-CAM approach \cite{GradCAM}, then mathematically,
\begin{equation}
\label{eq:Grad_cam}
\begin{aligned}
\mathbf{Y^c_{ij}} &= \mathbf{ReLU} \left (\sum\limits_k \alpha^c_k  A^k_{ij} \right )
\end{aligned}
\end{equation}
where,
\begin{equation}
\label{eq:alpha}
\begin{aligned}
\alpha^c_k &= \frac{1}{N}\sum\limits_i \sum\limits_j \frac{\partial y^c}{\partial A^k_{ij}}
\end{aligned}
\end{equation}
Here, $N$ is the normalization factor representing the number of pixels in the activation map $A^k$.
From equation \ref{eq:Grad_cam}, ReLU, applied to the linear combination of feature maps with the neuronal weights, preserves the features which have positive influence on the class of interest. When we average the gradients, we may get a positive or negative value for the neuronal importance weight $\alpha$.

Typically, $\frac{\partial y^c}{\partial A^k_{ij}}$ represents how the output score $y^c$ is affected upon changing the value of the activation unit $A^k_{ij}$. A positive value of $\frac{\partial y^c}{\partial A^k_{ij}}$ denotes that the output score $y^c$ increases when the value of the activation unit $A^k_{ij}$ increases (stronger activation). This implies that the unit $A_{ij}^k$ can capture certain features which the model relies upon in identifying class `$c$'. However, in the traditional Grad-CAM\cite{GradCAM} approach, as the importance weight, $\alpha^c_k$, is calculated by performing a global averaging operation on the output score-activation unit gradients, $\frac{\partial y^c}{\partial A^k_{ij}}$, instead of focusing on an individual unit $A^K_{ij}$, we are now looking at the entire activation map $A^k$ as a whole. Hence, in the event when a particular neuron importance weight $\alpha^c_k$ is negative, the entire activation map $A^k$ is assigned a negative value, weighted by $|\alpha^c_k|$ with the strongest activation $A^k_{ij}$ having the most negative value. This results in the loss of activation units that show a positive association with the output score. Furthermore, as each activation map $A^k$ is now scaled throughout positively or negatively by $\alpha^c_k$, the weighted combination of $\displaystyle\sum\limits_k\alpha^c_kA^k$ has a greater chance of being negative. This weighted combination, when passed through the ReLU unit, as described in equation \ref{eq:Grad_cam}, results in zero value class activation map being generated for the given class. Consequently, this results in inability to generate the activation maps for few samples as shown in Fig. \ref{Fig:Ch3_visualizations}(a). To overcome this limitation, we propose a slight modification in the Grad-CAM\cite{GradCAM} which will be discussed in the next section.

\subsection{Proposed Method of Visualization - Modified Grad-CAM}
In order to focus on the contributions of individual activation units towards a particular class, we propose a modified approach for generating the class activation map by making the use of the Hadamard product i.e. element-wise multiplication of gradient tensors with the respective feature maps. We call this method as modified Grad-CAM. This method avoids the possibility of zero activation maps and boosts the individual pixels' importance having a positive influence in favor of a particular class. 

Suppose  $\mathbf{X_{ij}^c}$ denote the ${ij}^{th}$ pixel for the one generated using proposed approach. Then, mathematically,
\begin{equation}
\label{eq:proposed}
\begin{aligned}
\mathbf{X^c_{ij}} &= \mathbf{ReLU} \left[\sum\limits_k \left(\frac{\partial y^c}{\partial A^k_{ij}} \cdot  A^k_{ij}\right) \right ]
\end{aligned}
\end{equation}
 Our model consists of only convolution layers followed by a linear classifier consisting of a dense layer as shown in Fig. \ref{Fig:Ch3_Baseline_model}. Consider the case for output activation map, $\mathbf{A^k}$
 \begin{equation}
 \begin{aligned}
\label{eq:weights_equal_gradiants}
y^c = \mathbf{(W^{ck})}^{T}\mathbf{A^k}+{b^c}
\implies \frac{\partial y^c}{\partial A^k_{ij}} = w^{ck}_{ij},\\
 \end{aligned}
\end{equation}
where, $y^c$ denotes the output logit for class $c$, $\mathbf{W^{ck}}$ is the weights connecting output logit for class $c$ to the input activation map $\mathbf{A^k}$ and $b^c$ is the bias term.
Therefore, from equation \ref{eq:weights_equal_gradiants}, we see that the gradient values i.e., $\frac{\partial y^c}{\partial A^k_{ij}}$, are equal to the weights connecting the softmax layer i.e. final layer and the activation maps of the penultimate layer. 
 
\begin{figure}[t]
    \centering
    \includegraphics[scale=0.3, trim={0 0cm 0 0}, clip]{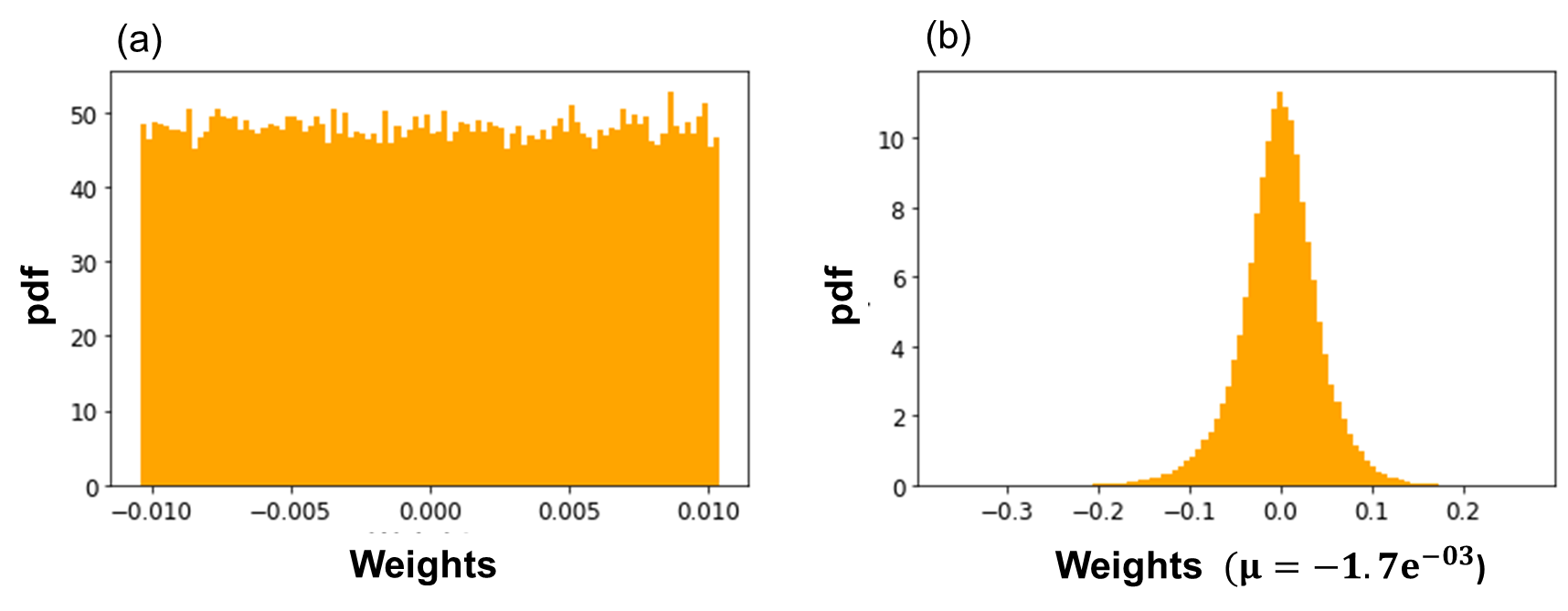}
    \caption{(a) Initial weight distribution, (b) Distribution of weights after training}
    \label{Fig:Ch3_weights}
\end{figure}

The weights in the final (dense) layer, $\mathbf{W} = \left\{\mathbf{W}^{ck}\right\},\,\,\forall\,\,c \in$ \{\# of classes\}$\,\,\& \,\,k \in$ \{\# of feature maps in penultimate layer\} are initialized from the uniform distribution as shown Fig. \ref{Fig:Ch3_weights}(a). From the experimentation, we have observed that these weights can be closely approximated by a normal distribution with mean $\mu =$ sample mean and variance $\sigma^2$ (say) as shown in Fig. \ref{Fig:Ch3_weights}(b), once the network is trained.  So, if $\mathbf{X_{ij}^c}$ denotes the ${ij}^{th}$ pixel for the generated class activation map using the Hadamard product and $\mathbf{Y_{ij}^c}$ denote the ${ij}^{th}$ pixel for the one generated using the conventional Grad-CAM approach \cite{GradCAM}, then assuming the weights to be independent and identically distributed (I.I.D.), one can write:
\begin{equation}
\begin{aligned}
\mathbf{X^c_{ij}}&= \sum\limits_k w^{ck}_{ij}A^k_{ij} \sim \mathcal{N}\left(\underbrace{\mu\sum\limits_k A^k_{ij}}_{\mu_1}, \underbrace{\sigma^2\sum\limits_k (A^k_{ij})^2}_{\sigma_1^2}\right)\\
\end{aligned}
\end{equation}

\begin{equation}
\begin{aligned}
\mathbf{Y^c_{ij}} &= \sum\limits_k\left(\sum\limits_{l,m}w^{ck}_{lm}\right)A^k_{ij} \sim \mathcal{N}\left(\underbrace{n^2\mu\sum\limits_k A^k_{ij}}_{n^2\mu_1}, \underbrace{n^2\sigma^2\sum\limits_k (A^k_{ij})^2}_{n^2\sigma_1^2}\right)
\end{aligned}
\end{equation}

Here $A^k \in \mathbb{R}^{n\times n}$. Thus, we have $X^c_{ij} \sim \mathcal{N}\left( \mu_1,\sigma_1^2\right)$ and  $Y^c_{ij} \sim \mathcal{N}\left( n^2\mu_1, n^2\sigma_1^2\right)$

\begin{equation}
\label{eq:propability_prop}
\begin{aligned}
\text{Pr}\left({\mathbf{X^c_{ij}}} > 0 \right) &=  \text{Pr} \left(\frac{\mathbf{X^c_{ij}} - \mu_1}{\sigma_1} > \frac{-\mu_1}{\sigma_1} \right)
\end{aligned}
\end{equation}

\begin{equation}
\label{eq:propability_grad}
\begin{aligned}
\text{Pr}\left({\mathbf{Y^c_{ij}}} > 0 \right) &=  \text{Pr} \left(\frac{\mathbf{Y^c_{ij}} - n^2\mu_1}{n\sigma_1} > \frac{-n\mu_1}{\sigma_1} \right)
\end{aligned}
\end{equation}

In such a case, if $\mu < 0$ (as shown in Fig. \ref{Fig:Ch3_weights}(b)) then we can conclude that, 
\begin{equation}
    \text{Pr}\left(\mathbf{X}^c_{ij} > 0\right) > \text{Pr}\left(\mathbf{Y}^c_{ij} > 0\right)
\end{equation}
Consequently, the classical Grad-CAM approach defined in equation \ref{eq:Grad_cam} is more likely to generate activation maps with 'zero' activations (because of \textit{ReLU}) as compared to proposed approach as per equation \ref{eq:proposed}. Assuming the weights to be I.I.D and normally distributed, we have proven that the probability of generating visualization maps for Modified Grad-CAM is more than classical Grad-CAM\cite{GradCAM}. Through experimentation, we have found that the weights initialized from the uniform distribution shown in Fig. \ref{Fig:Ch3_weights}(a) are transformed to have distribution with mean $\mu=-1.7e^{-3}$, and it can be approximated by Gaussian distribution. This corroborates that our assumption of Gaussianity.

\subsection{Comparing the visual explanations by humans and model}
Both fixation maps by humans and visualization maps by model generate the human-understandable explanations for their respective decisions, and hence we call these maps `visual explanations’. The qualitative comparison between the visualization maps and fixation maps has revealed an interesting correlate. The samples that model and humans correctly classify have shared similar informative regions. In other words, both humans and model have focused on similar regions while making a decision in favor of a correct class. On the contrary, there is a mismatch between the focused regions for samples misclassified by the model. We hypothesize that when the model can not find discriminating character regions and relies on other character regions (like artistic elements attached to the essential character structure) for learning features for classification, the character gets misclassified. In other words, the misguided focus on the different character regions other than discriminating regions results in misclassificaton of that sample by the model. Assuming that human fixation captures discriminating character regions in character image, it is interesting to check `whether the model's performance improve if we shift the model's focus on regions fixated by humans? Will the misclassified samples be correctly classified by such reorientation of focus?' To answer these questions, we decide to re-align the model's focus to that of humans via an attention mechanism.

\subsection{Visual Explanations Guided Attention Model (VEGAM)}
\begin{figure*}[t]
    \centering
    \includegraphics[scale=0.46, trim={0 0cm 0 0}, clip]{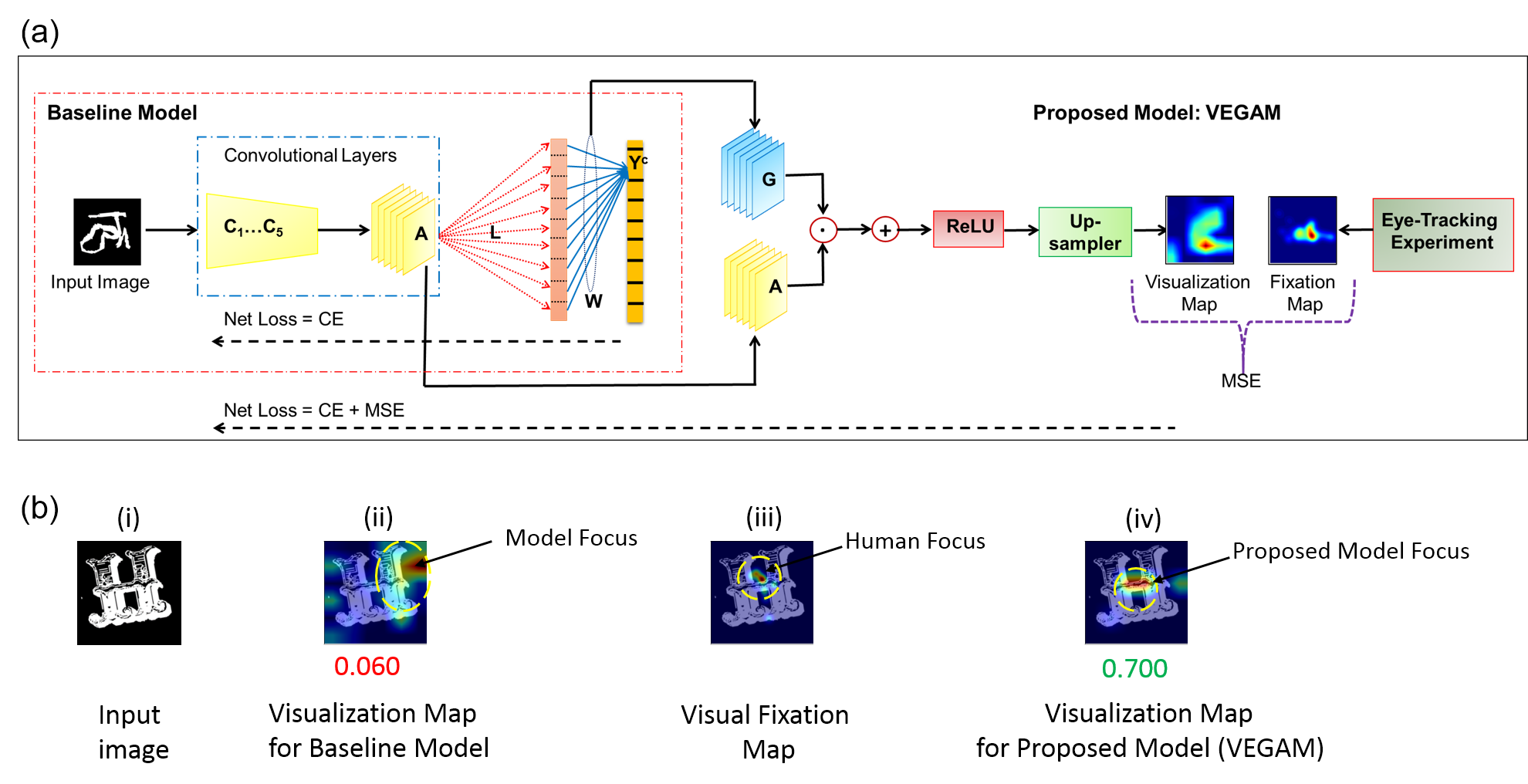}
    \caption{(a) Proposed attention model, VEGAM uses CE and MSE loss for training the model. `A' is the forward activation maps of the last convolution layer (i.e. $C_6$) and `G' is the gradients of class score w.r.t. A. `W' represents the weights connecting the last layer and activation maps which are also equal to the gradients `G'. (b) (i) Input image, (ii) The baseline model misclassifies the sample, and it's focus is encircled yellow. The confidence corresponding to the true class is 0.06. (iii) Visual fixation map. (iv) Proposed model `VEGAM' shifts the focus to regions fixated by humans, and it results in correct classification of that sample with increased confidence (i.e. 0.700) corresponding to the true class.}
    \label{Fig:Ch3_VEGAM}
\end{figure*}

In this attention model, we aim to make model focus on the similar regions as fixated by humans. We can achieve this by minimizing the mean square error (MSE) between the model-generated heat maps of explanations (visualization map) and the heat map of visual fixation (fixation map). The hope of improving the model performance by minimizing the error between two maps motivates us to use the fixation maps to supervise the model. The entire training process is formulated as an optimization problem with two losses viz. classification loss (i.e. cross-entropy (CE) loss) and the mean squared error (MSE) loss. Here, we have a single model with two different training strategies. For simplicity, we name the the model which is trained with images as inputs and CE as loss function as our baseline model; the model that uses fixation map as an additional supervisory input which is trained with CE and MSE loss, is named as `Visual Explanations Guided Attention Model (VEGAM)'. The schematic of the proposed model is shown in Fig. \ref{Fig:Ch3_VEGAM} (a).

\subsubsection{Training the model}
We have used the character images as input and cross-entropy (CE) loss function for training our baseline model. In VEGAM, we have used mean squared error(MSE) and CE as losses for training the model. Therefore, we are referring to these models as two separate models. The visualization maps are resized to that of visual fixation maps using bilinear interpolation, which we use to calculate MSE. The net loss for `VEGAM' is the sum of CE and MSE loss which are weighted by some hyper-parameter. We have used Adam optimizer with learning rate 0.001 for training and batch size is 64. We have used the regularization techniques (such as batch-normalization\cite{batchnorm} and dropout\cite{Dropout}) and early stopping to avoid the problem of over-fitting. The implementation and the experimentation of the network is carried out using PyTorch. 

From equation \ref{eq:weights_equal_gradiants}, it is evident that the gradients of the maximum logit (predicted output) w.r.t. activation maps of the penultimate layer are nothing but the weights connecting the last linear layer. As a result, these randomly initialized weights can be used to generate the visualization maps during training for the first epoch. Such visualization maps are compared with fixation maps via MSE.  Initial visualization maps generated from random weights may not be as good in indicating the highly discriminating regions compared to the maps for a fully-trained model. However, in subsequent training epochs, the weights will be updated which will highlight the important image regions. Generating the visualization maps during training allows us to train the model to focus on the same regions as that of humans and optimize both CE and MSE losses simultaneously.

\subsection{Model evaluation}
We have used classification accuracy as the metric to compare the performance of the two models. It denotes the proportion of correctly classified samples from the total number of samples. In addition, we have compared predictions of the two models using McNemar test. We have used the confidence towards the actual label for model comparison.  This confidence is nothing but the output of the softmax layer (i.e. probability) corresponding to the true class label.

\subsection{Aligning model’s focus improves performance}
\begin{figure*}[t]
    \centering
    \includegraphics[scale=0.35, trim={0 0cm 0 0}, clip]{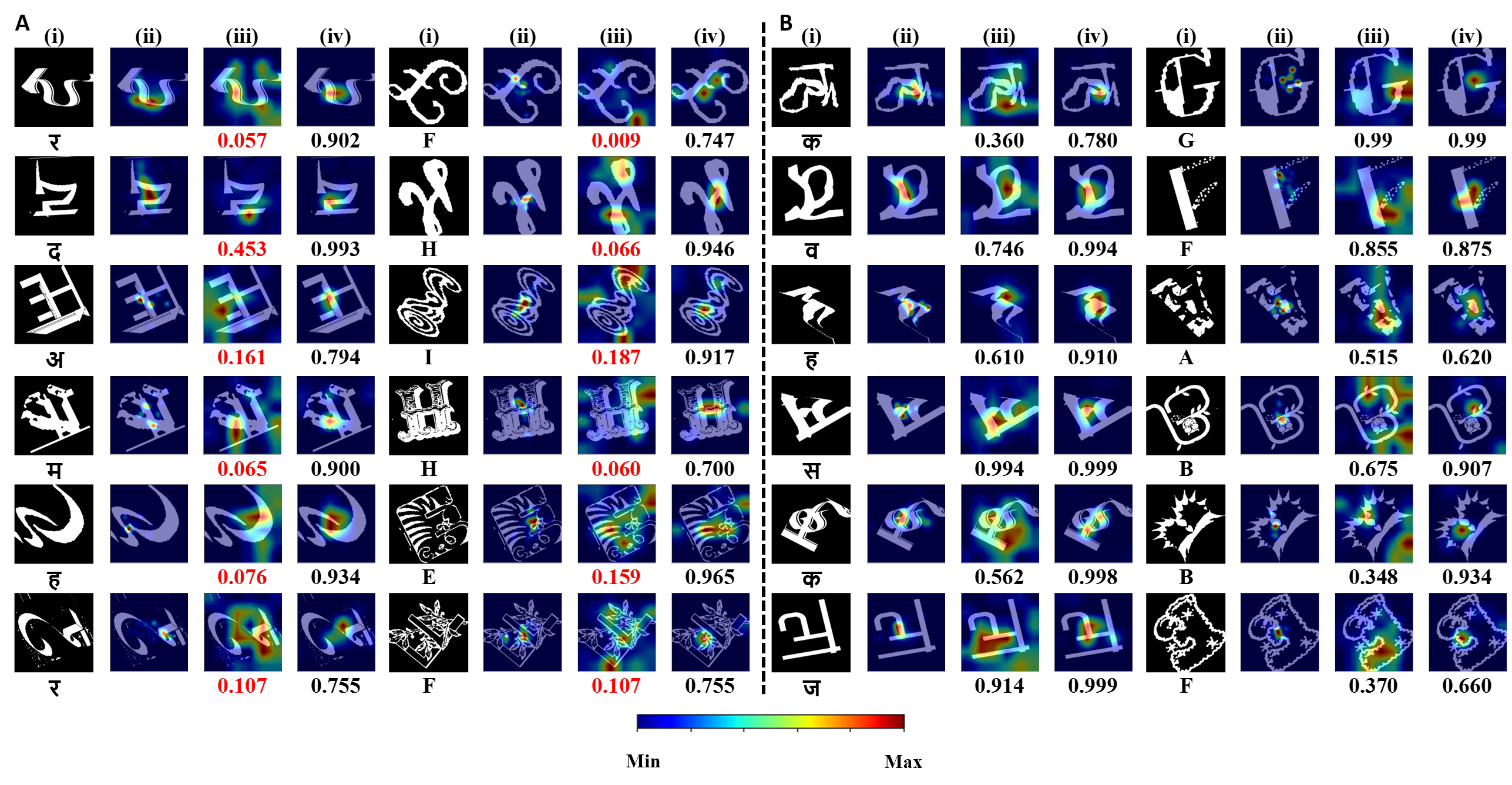}
    \caption{(i) Input image, (ii) Fixation map, (iii) Visualization maps generated for baseline model, (iv) Visualization maps generated for the proposed model. The numbers below the images indicate the confidence corresponding to actual class-label. (A) Samples misclassified by baseline model are getting correctly classified by our proposed model, (B) Samples correctly classified by baseline model remain correctly classified with an increase in confidence corresponding to the true class label}
    \label{fig:Fig_4}
\end{figure*}

To verify whether the VEGAM model has shifted it’s focus towards the similar regions focused by humans, we have generated the visualization maps for the baseline model and proposed the attention model. Fig. \ref{Fig:Ch3_VEGAM} (b) (ii) shows that the baseline model fails to focus on the relevant character for extracting features, because of which it misclassifies the character. Using the proposed attention model (VEGAM), we can observe the change in regions used for feature extraction as shown in Fig. \ref{Fig:Ch3_VEGAM} (iv). When we train the model to shift it's focus on regions elicited by fixation maps, the misclassified samples get correctly classified. Using this approach, we have observed an improvement in the model's performance by $\sim$ 2 to 3\%  for both Devanagari and English (Latin) characters over their respective baseline counterparts as shown in table \ref{tab:Accuracy_table} and this performance improvement is statistically significant. We observe that both humans and the model focus on similar character regions for the samples that are correctly classified by the baseline model and humans. 
To understand the consequences of proposed supervision by fixation maps on such samples, we have calculated the probability score corresponding to the true class label for the character image. We can also term this as confidence towards the true class label. This score is shown in Fig. \ref{fig:Fig_4}(A) and Fig. \ref{fig:Fig_4}(B) for every character. It is evident from the results that the use of the proposed attention mechanism increases the confidence for the actual class label. Thus, we can conclude that human eye-fixations are helping the model perform better by focusing on the informative regions.

\begin{table}[t]
    \centering
    \caption{Recognition performance on testing samples of Devanagari and Latin dataset}
    \label{tab:Accuracy_table}
    \begin{tabular}{ccc}
    \hline
    \textbf{Network Description} & \textbf{Devanagari Dataset} & \textbf{Latin Dataset}\\
    \hline
    \textbf{Baseline Model}  & 80.0\% & 78.0\%\\
    \textbf{VEGAM} & 82.67\% & 81.20\%\\
    \hline
    \end{tabular}
\end{table}

\section{Discussions}
Character recognition has been a topic of interest in the pattern recognition community for few decades. Intensive research has been carried out for building an intelligent recognition model by extracting the best possible features.  As a result, we have witnessed the transition from the selection of carefully engineered, handcrafted features\cite{basu2009hierarchical,sarkhel2016multi,joshi2005machine} to automatic feature extraction using deep neural networks\cite{mehrotra2013unconstrained,roy2017handwritten} for developing character recognition models. In the case of character recognition, the fundamental question is `how is a character recognized?' If we can understand the process of character recognition by DNNs and humans, we can build a better recognition model. 

Investigation of visual processing may help in unlocking the process of recognition. In humans, visual information processing starts from V1 and ends at the inferior temporal (IT) cortex where the recognition of a particular category happens\cite{tang2018recurrent, dicarlo2012does}. The actual algorithms used for information processing can be decoded by monitoring the activity of thousands of interconnected neurons simultaneously\cite{Goodfellow_book}. However, it might be challenging to precisely acquire information about neuronal activity with non-invasive brain investigation techniques like EEG, fMRI, and MEG. On similar lines, understanding internal functioning in DNNs seems difficult due to the involvement of highly complex non-linear mapping between input and output\cite{Yann}. Ultimately, the process of character recognition remains an open problem.

Character perception is defined as the way in which we construct the characters from visual features or regions\cite{finkbeiner2009letter}. These visual features determine the recognition of that character. Recently, it has been shown that humans rely significantly on specific visual features, whereas most of the machine vision algorithms do not consider such dependence on specific regions for recognition\cite{ullman2016atoms}. The results from our eye-tracking experiments suggest such reliance on specific character regions for it's recognition. The selective focus on different character regions inferred from fixation maps governs the observer's decision. When humans focus on the diagnostic character regions, the character is getting correctly recognized. Similar to humans, DNNs seem to rely on specific character regions which can be inferred from visualization maps. Our visualization method highlights the important character regions considered by the model for making a decision in favor of a class.

The qualitative comparison between fixation maps and visualization maps demonstrates an interesting correlate of congruence for correctly classified samples. Both humans and the model consider similar character regions in case of the correctly classified samples. We can say that when the information uptake by the model is from similar regions as that of humans, the corresponding character is correctly classified. Although we can not explicitly comment on the internal functioning of the model or the human brain, the similarity in the highlighted regions for correctly classified samples may signify that the model is able to capture important character regions as that of humans. However, when it fails to do so, the sample gets misclassified. This argument is backed by the results showing improvement in model performance and a substantial increase in the true class confidence when we make the model shift it's focus to human fixated regions. Such alignment of focus is achieved in our proposed model `VEGAM' where we have used fixation maps as the supervisory input. 

Although there is a significant increase in the recognition performance of `VEGAM', the main objective of this paper is to consider the importance of understanding the underlying process of recognition. Nowadays, DNNs are approaching human-level performance in recognition, and various other computer vision tasks through network engineering wherein the factors such as depth, width, and cardinality of the model are varied\cite{woo2018cbam}. In addition to network engineering, attention mechanisms are also incorporated in the models\cite{Mnih,xu2015show,woo2018cbam, ralekar2019intelligent,ralekar2021collaborative}. In most of the cases, these methods requires additional parameters and hyper-parameters to be tuned for the best possible performance. The proposed model `VEGAM' does not require additional parameters and has considered visual attention.  

The visual attention derived from eye-fixations has been used in some studies\cite{ralekar2019intelligent,ralekar2021collaborative}. However, these models hardly consider the explanability. As we are more interested in generating the explanations for a classifier's decision and using those explanations for better recognition, we have chosen an architecture comprising convolutional layers followed by a linear classifier. Perhaps, use of a non-linear classifier may result in better recognition performance. Hence, in the future, we will try to explore the possibility of using non-linear classifier while developing attention model based on the visual explanations provided by humans and the model.

\section{Conclusions}
In this paper, we have observed that humans and DNNs rely on selective information uptake while classifying a character. This information selection strategy can be understood by visualizing the important, informative character regions that ultimately govern the decision of humans and DNNs. Our results demonstrate that the proposed method of generating explanations for DNN is better compared to conventional Grad-CAM visualization. Our approach of deriving visual explanations from eye-fixation and visualization maps draws an interesting parallel between the processes undertaken to recognize character by humans and DNNs. Eye-fixation maps reveal that humans focus on the highly diagnostic and discriminating character regions. When used as a supervisory input, such eye-fixation maps has the potential of making artificial systems of character recognition more accurate. The experimental results from this research attest that superior machine vision performance can be achieved when augmented with cues from cognitive experiments. One way to bridge the gap between machine and human intelligence is to design the artificial systems using cues derived from their natural counterpart, such as through cognitive experiments on human participants. This, henceforth,  motivates a need for substantial efforts in exploring various aspects of the human brain and develop the systems based on the findings.



\section*{Acknowledgment}
We would like to thank all the participants participated in the study. We want to thank the Visvesvaraya PhD Scheme/DIC/Meity  for providing financial assistance to the first author with Unique Awardee Number: MEITY-PHD-589, to carry out research. Special thanks to Prof. Pawan Sinha from MIT, USA and Prof. S. D. Joshi from IIT Delhi for their valuable suggestions.

\ifCLASSOPTIONcaptionsoff
  \newpage
\fi



%


\bibliography{refs.bib}
\bibliographystyle{IEEEtran}

%








\end{document}